\documentclass[letterpaper, 10 pt, conference]{ieeeconf}  

\IEEEoverridecommandlockouts                              

\usepackage{graphicx} 
\usepackage{amsmath} 
\usepackage{amssymb}  

\usepackage[sorting = none, backend = bibtex, style=numeric-comp]{biblatex}
\usepackage{ifluatex,ifxetex} 
\ifluatex
  \usepackage{fontspec} 
\else\ifxetex
  \usepackage{fontspec} 
\else 
  \usepackage[T1]{fontenc}
  \usepackage[utf8]{inputenc} 
\fi\fi

\newcommand{\dd}{\mathop{}\!\mathrm{d}}

\title{\LARGE \bf
PCBot: a Minimalist Robot Designed for Swarm Applications
}

\author{Jingxian Wang$^{1}$ and Michael Rubenstein$^{1}$
\thanks{$^{1}$Jingxian Wang and Michael Rubenstein are with the Center for Robotics and Biosystems,
        Northwestern University, Evanston, IL 60601, USA
        {\tt\small jingxianwang2026@u.northwestern.edu, rubenstein@northwestern.edu}}%
}

\begin{document}

\maketitle
\thispagestyle{empty}
\pagestyle{empty}

\begin{abstract}

Complexity, cost, and power requirements for the actuation of individual robots can play a large factor in limiting the size of robotic swarms. Here we present PCBot, a minimalist robot that can precisely move on an orbital shake table using a bi-stable solenoid actuator built directly into its PCB. This allows the actuator to be built as part of the automated PCB manufacturing process, greatly reducing the impact it has on manual assembly. Thanks to this novel actuator design, PCBot has merely five major components and can be assembled in under 20 seconds, potentially enabling them to be easily mass-manufactured. Here we present the electro-magnetic and mechanical design of PCBot. Additionally, a prototype robot is used to demonstrate its ability to move in a straight line as well as follow given paths. 

\end{abstract}

\section{INTRODUCTION}
There are many envisioned applications of swarm robots, from distributed mapping and exploration \cite{fox2006distributed, chiu2009tentacles}, modular self-reconfigurable robots \cite{gilpin2010robot,swissler2018fireant,yim2007modular}, robotic warehouses \cite{wurman2008coordinating}, shape formation \cite{gauci2018programmable}, etc.  Often, these swarms' capabilities improve with increased numbers of robots. For example, a swarm with more robots can push heavier objects or move them faster in collective transport tasks \cite{gross2009towards,rubenstein2013collective}, or can approximate a desired shape in more detail in shape formation tasks \cite{gilpin2010robot}. 

The size of swarms in use is typically constrained by the robots' complexity, cost, and ease of manufacturing. To create larger swarms, designers are often forced to make them as simple, low cost, and easy to manually assemble as possible. Swarms of more complex robots such as SWARM-BOT \cite{mondada2003swarm} or e-puck \cite{mondada2009puck} have been limited to under 100 robots.  There were attempts to simplify swarm robots in the past, such as Kilobot \cite{rubenstein2014programmable}, but even then, five minutes of manual assembly per robot limited the number of robots in a swarm to approximately 1000.  On the contrary, biology frequently makes use of swarms and can have swarms that number in the 100,000 or millions of individuals such as those found in ant colonies, bird murmurations, or schooling fish.   

One way to simplify the robots and ease their manufacturing is to rely on more automated manufacturing techniques for their assembly.  One approach is to attach electrical components to a sheet that then self-folds to create the robot \cite{felton2014method, nisser2016feedback}, though robots created with this technique so far have been quite limited.  Another approach is to take advantage of advances in printed circuit board (PCB) fabrication and component placement to build a robot almost entirely out of a PCB, such as the HoverBots \cite{10.3389/frobt.2017.00055}.  While Hoverbot partially inspired PCBot, it had limited battery life and can only work on a complex air-table with embedded magnets.

\begin{figure}[t]
\centering
\includegraphics[width=\linewidth]{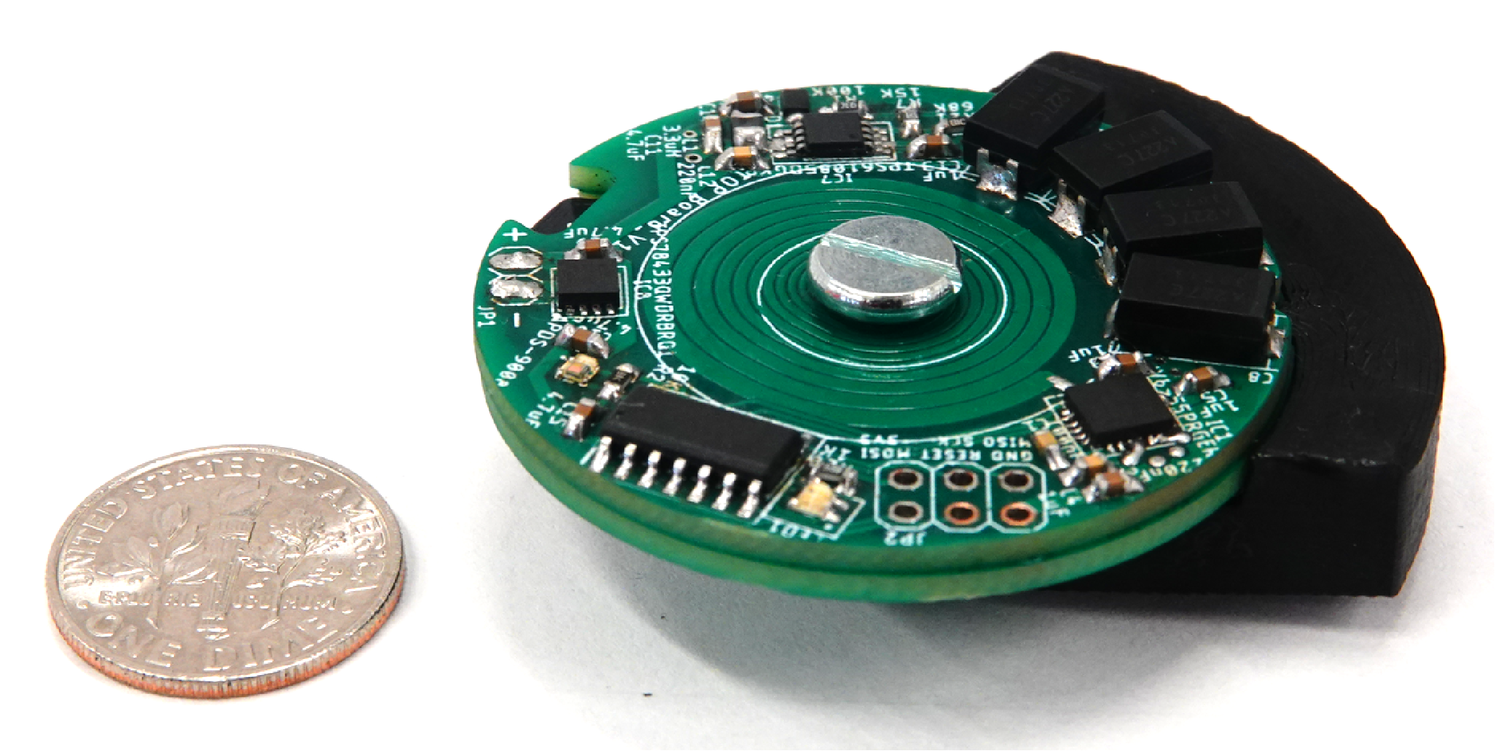}
\caption{A photo of PCBot, a robot that
uses a PCB-based actuator, and is designed for ease of manufacturing. PCBot's size is $48\times42\times14\,\text{mm}$ and weighs $18\,\text{g}$.}
\vspace{-5mm}
\label{spinbotphoto}
\end{figure}

The number of actuators, as well as their complexity, also contributes a lot to a robot's overall complexity. Traditional robots use multiple motors and gearboxes in combination with wheels \cite{wurman2008coordinating,mondada2009puck} or propellers \cite{GREGORY2020e00105,6301066} to move the robot. These driving mechanisms limit the robot's cost to hundreds of dollars, and part count to tens. In recent years, researchers have tried multiple ways to reduce the number and complexity of actuators, resulting in robots simpler in design and easier to manufacture. Kilobot \cite{rubenstein2014programmable}, I-SWARM \cite{4107855}, and bristle bots \cite{8206295,fortunic2017design} replace traditional motors with vibration motors to eliminate the need of wheels and gearboxes. These robots usually have around 10 components and are a lot simpler than traditional robots, but the stochastic nature of vibration makes their movement unpredictable. On the other hand, Piccolissimo \cite{piccoli2017piccolissimo}, Monospinner \cite{zhang2016controllable}, and 1STAR \cite{6907226} seek to reduce the number of actuators used to drive the robot. These robots feature only one actuator, usually directly 3D-printed body, and less than 10 components.

Robots with extremely simple actuators often rely on added complexity to the external environment. For example, HoverBots \cite{10.3389/frobt.2017.00055} are composed of only two major components: a battery and a PCB, and can move themselves on a special `levitation–magnet table' using magnetic force between actuators integrated into the PCB and the table. However, the `levitation–magnet table' is embedded with arrays of magnets and is complicated to manufacture and scale up. A simple robot described in \cite{wang2016autonomous} and Spinbot \cite{saloutos2019spinbot} utilize a planar table moving in an orbital manner to simplify the robot's locomotion. These two robots use electro-permanent magnets (EPMs) to control their attachment and detachment from the orbital shake table. EPMs used in these robots are relatively time-consuming to produce \cite{ceron2019towards}, but give them the ability to move in any 2D direction on the table. While these robots are more complicated, the table they operate on is only a steel plate driven by several motors, and is simple to manufacture and scale up.

Here we present a robot called PCBot that builds on Spinbot \cite{saloutos2019spinbot} and HoverBot \cite{10.3389/frobt.2017.00055} to further reduce the complexity of a robot system designed for swarm applications.  This robot can precisely move on an orbital shake table using a novel PCB actuator inspired by Hoverbot.

\begin{figure}[hbtp]
\centering
\vspace{-3.5mm}
\includegraphics[clip,trim=0cm 1.5cm 0 1cm,width=\linewidth]{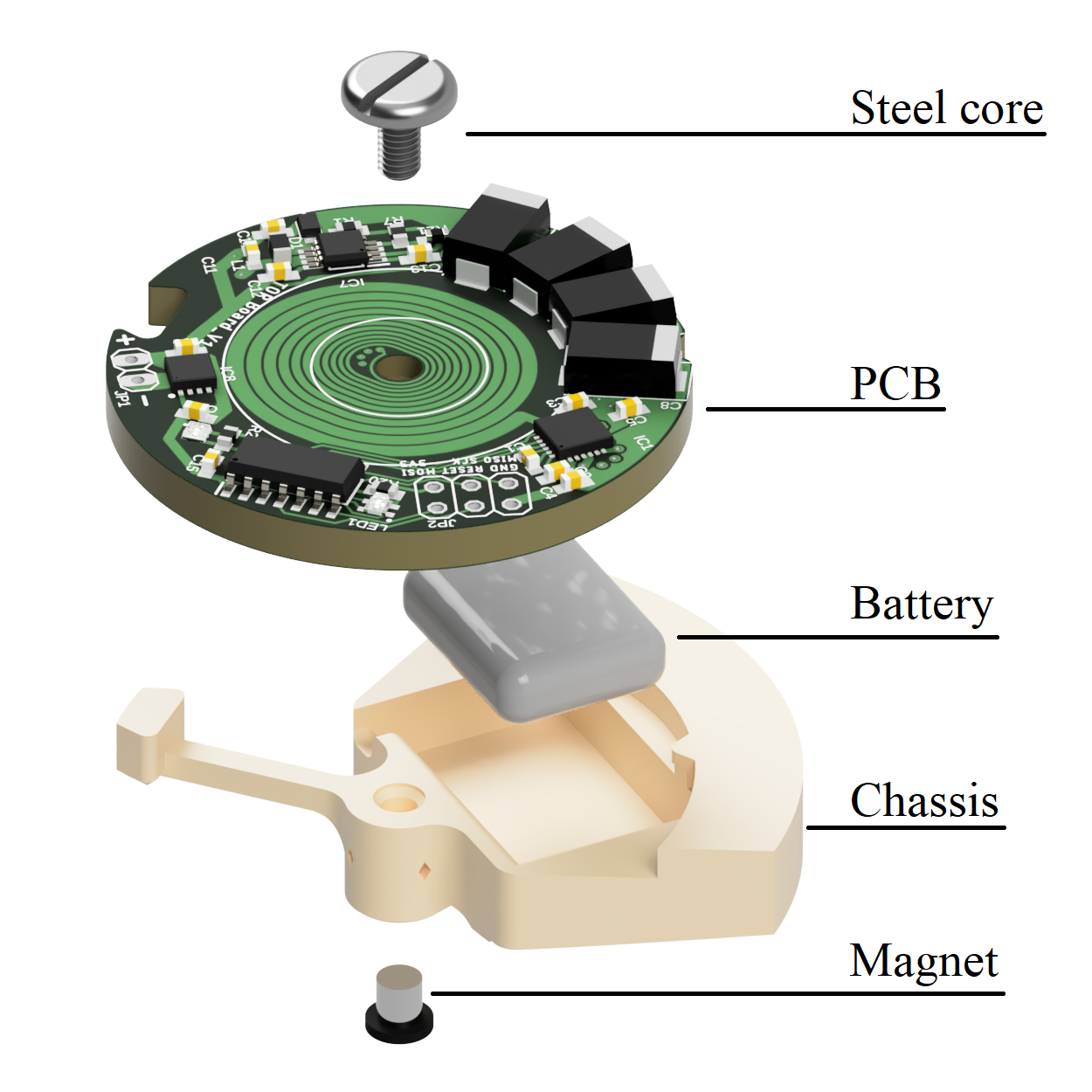}
\caption{An exploded view of PCBot. PCBot is comprised of merely five major components: a 3D printed chassis, a PCB, a battery, a steel core which is just a screw, and a magnet with rubber on one side. All these components are off-the-shelf or mass-manufacturable and can be assembled in under 20 seconds.}
\vspace{-2mm}
\label{explode}
\end{figure}

\section{Robot Design}\label{design}

\subsection{Overall concept}\label{overall-concept}

\begin{figure}[htbp]
\centering
\includegraphics[width=\linewidth]{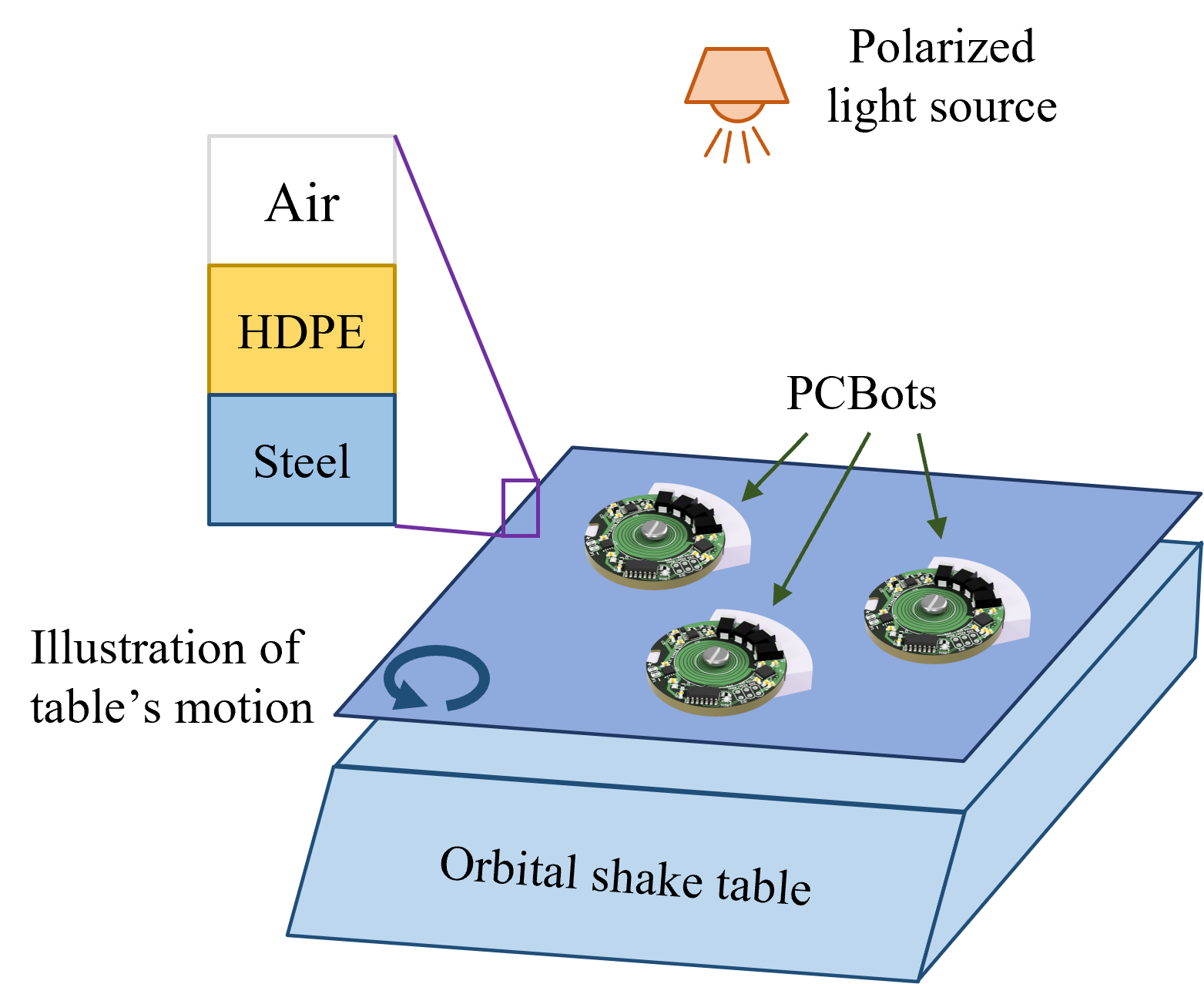}
\caption{The setup of the test environment. A polarized light source hangs from the ceiling to provide robots with global orientation sensing. The orbital shake table surface is made of steel and is coated with a layer of HDPE sheet. PCBots operate on the surface of the orbital shake table.}
\label{setup}
\end{figure}

\begin{figure}[htbp]
\centering
\includegraphics[clip,trim=0cm 0.3cm 0cm 0cm,width=\linewidth]{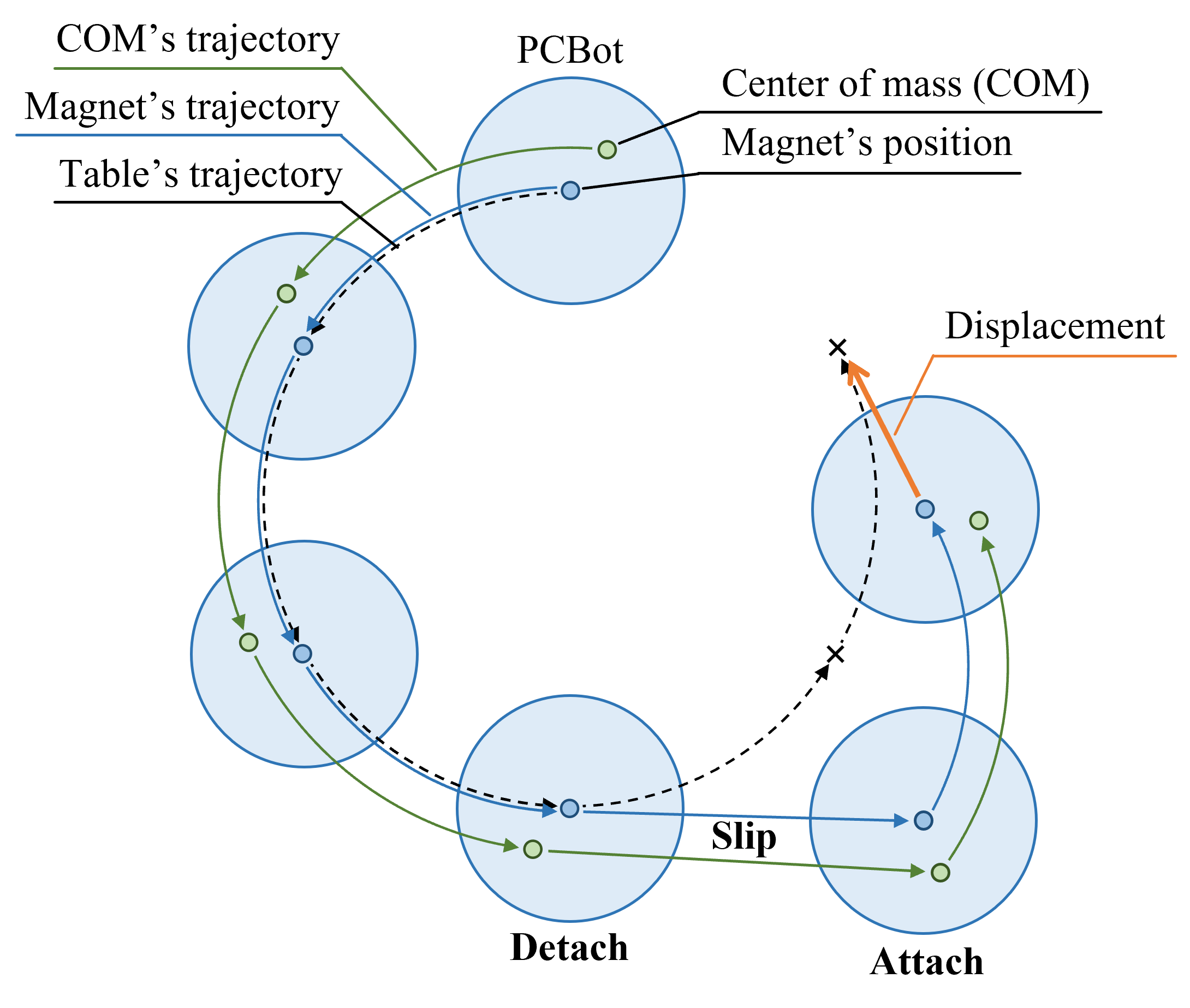}
\caption{A simplified diagram showing how the robot moves. In the `attached' state, as shown in the left part of the figure, the magnet sticks to the table and PCBot revolves around the magnet. When PCBot detaches, the magnet no longer touches the table. Driven by inertia, PCBot maintains its momentum, thus follows a straight path and slides on the table, as shown in the bottom part of the figure. After switching back to `attached' state, friction force quickly brings PCBot to the same speed as the table and keeps them relatively static, as shown in the right part of the figure. In each actuation cycle, PCBot moves relative to the table, as indicated by the orange arrow. This figure is not drawn to scale.}
\vspace{-3mm}
\label{movement}
\end{figure}

PCBot, see Figs. \ref{spinbotphoto}, \ref{explode}, sets out to further simplify the manufacturing of a swarm robot by introducing a novel bi-stable solenoid actuator built using coil embedded directly into the robot's PCB. PCBot operates on a horizontal orbital shake table, shown in Fig. \ref{setup}. As shown in previous work \cite{wang2016autonomous,saloutos2019spinbot}, individuals can move in an arbitrary 2-D direction relative to the table surface by simply controlling when it sticks (attaches) and slips (detaches) on the table, as shown in Fig. \ref{movement}. While using this single orbital shake table increases the complexity of the setup, it decreases the overall swarm complexity because it allows the individual  robot design to be much simpler.  

In addition to sliding over the table surface, PCBot also passively spins around its center in synchronization with the movement of the table when operating. This feature enables PCBot to track the movement of the table using its polarized light sensor, saving the need for gyroscopes. In future generations, the rotation behavior will enable PCBots to sense distance and bearing to neighbors as well as communicate with them using mechanisms similar to those presented in \cite{saloutos2019spinbot}.

\begin{figure}[htbp]
\centering
\vspace{-2mm}
\includegraphics[clip,trim=1.5cm 4.7cm 1.5cm 5cm,width=\linewidth]{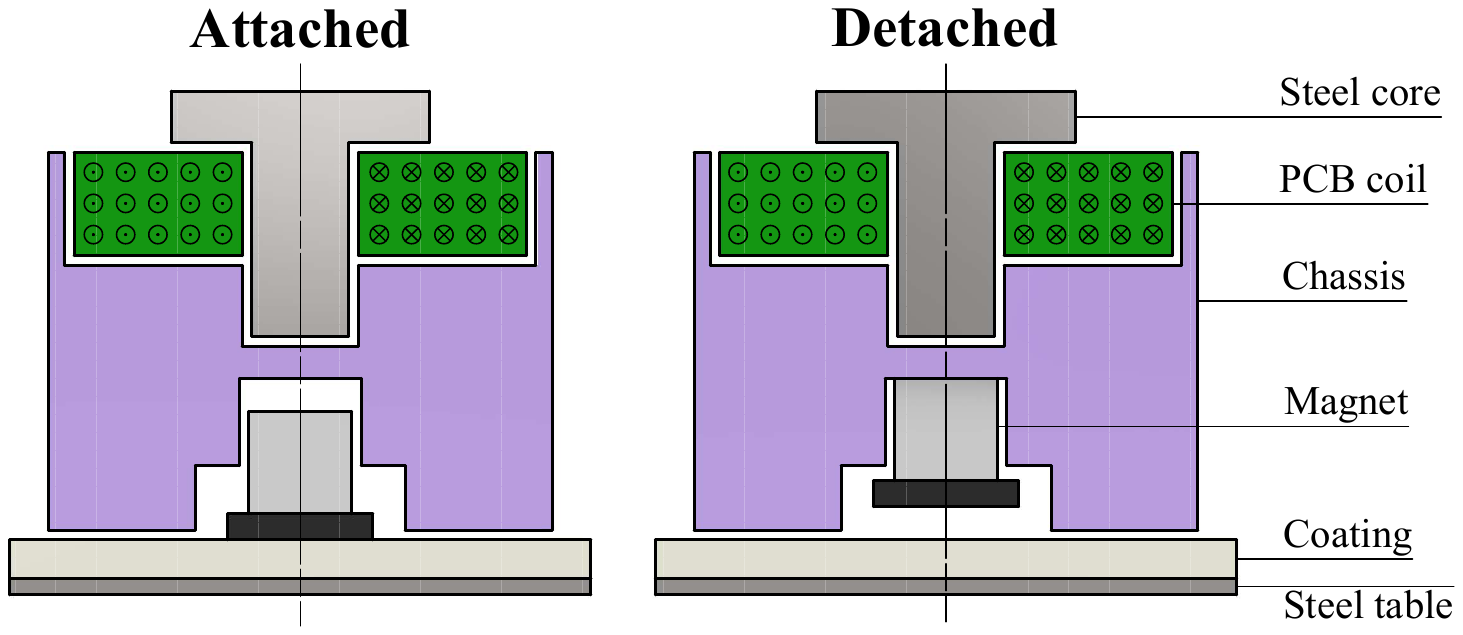}
\caption{A simplified cross-section view of the electro-magnetic subsystem of PCBot. This figure shows the difference between the magnet's position in the `attached' and `detached' states. In `attached' state, the magnet sticks to the surface, and in `detached' state, the magnet sticks to the spacer on the chassis close to the steel core. }
\vspace{-3mm}
\label{magnetsys}
\end{figure}

For PCBot's only actuator, we designed a novel bi-stable solenoid to quickly control the friction coefficient between the robot and the table surface. The bi-stable solenoid, as shown in Fig. \ref{magnetsys}, consists of a coil created directly by traces in the robot's multi-layer PCB and a coil-driven magnet that can move up and down in the chassis and switch between sticking to the orbital shake table (called `attached' state) or sticking to the spacer on 3D printed chassis close to the steel core (called `detached' state). The concept of creating a coil directly in a PCB is inspired by \cite{10.3389/frobt.2017.00055}. This design allows us to build an actuator without the manufacturing complexity of wire-wound coils or electro-permanent magnets \cite{wilson2020scalable,saloutos2019spinbot} and significantly reduces the major components required to build a single robot to merely five: a 3D printed chassis, a PCB, a battery, a steel core (a simple screw), and a magnet with rubber on one side. This consequently reduced the time required for manual assembly from tens of minutes or hours seen with most other robots, down to under 20 seconds for PCBot.

\subsection{Theory of operation}\label{theory-of-operation}

The overall environment for this robot consists of an arbitrary number of PCBots all moving on an orbital shake table. In addition, there is a polarized light source shining down on the robots, see Fig. \ref{setup}. Each PCBot is equipped with an ATTINY84 microcontroller, a $70\,\text{mAh}$ rechargeable lithium-ion battery, and a single actuation mechanism consisting of a bi-stable solenoid and its associated electronics. Furthermore, to sense its orientation in the environment as it spins around its magnet, PCBot uses an upward-facing light sensor covered with a polarizing filter. As a debugging aid, there is also an RGB LED to provide visual feedback to the operator. 

\begin{figure}[htbp]
\centering
\includegraphics[width=0.9\linewidth]{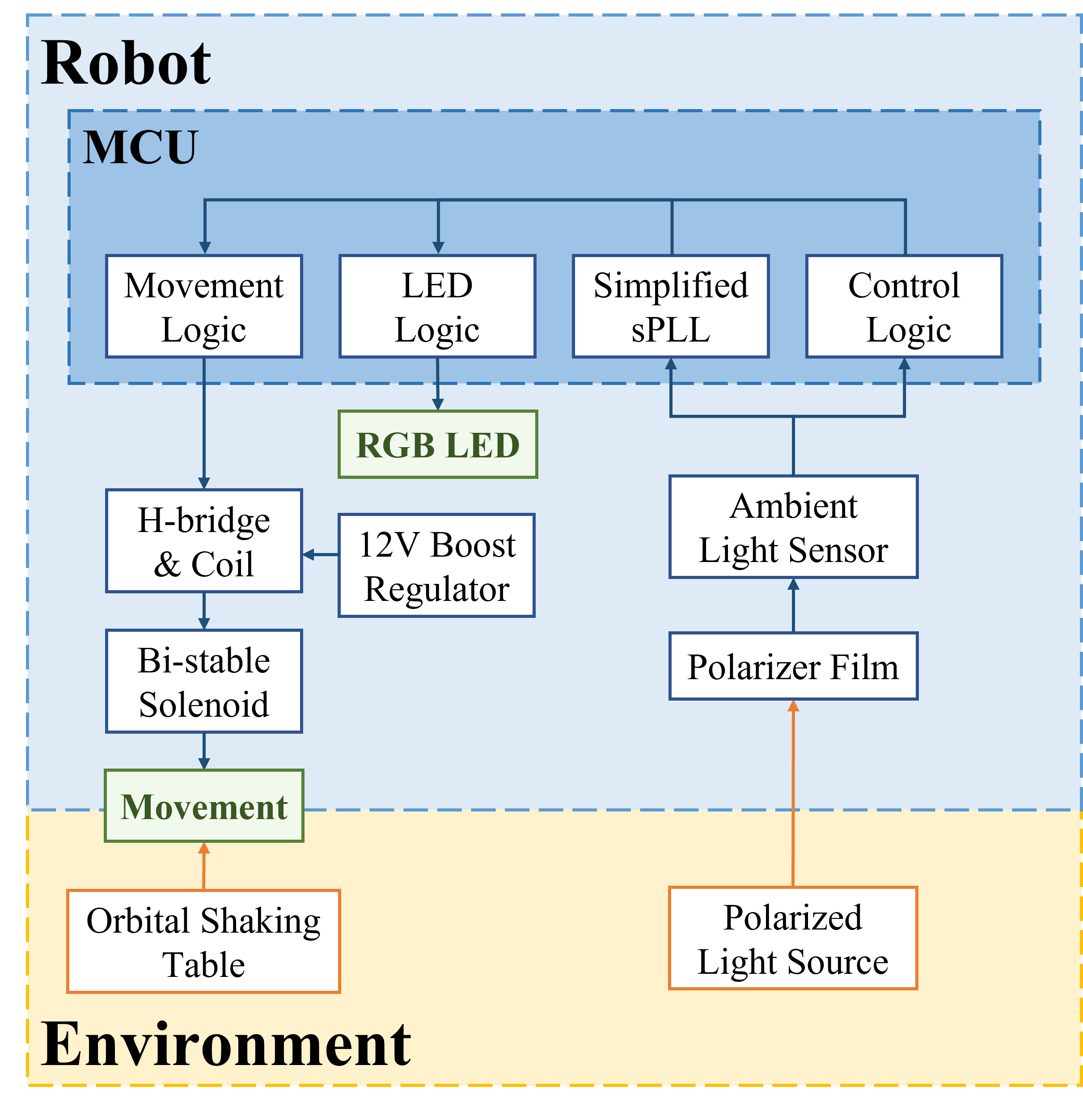}
\caption{Software and hardware architecture for the PCBot and its operating environment.}
\vspace{-3mm}
\label{schematic}
\end{figure}

At the center of PCBot is its bi-stable solenoid actuator which allows it to attach and detach from the table surface. The actuation coil is built in the multi-layer PCB and generates a magnetic field that moves the Neodymium magnet up or down in the 3D printed chassis, as shown in Fig. \ref{magnetsys}. A boost regulator and an H-bridge are used to selectively drive current through this coil. When no current is applied on the coil, the magnet is bi-stable: it could stay in either `attached' or `detached' state. In the `attached' state, there is a strong attraction force between the magnet and the steel beneath the table surface. Furthermore, there is a high friction coefficient between the rubber and the table surface. These two factors prevent the magnet from slipping, and the whole robot rotates around the magnet due to an offset center of mass. However, in the `detached' state, the chassis which has a low friction coefficient is the only part of the robot that is contacting the table, and the friction is not enough to allow the robot to continue moving with the table's movement. Thus, the robot slips and moves according to Newton's first law. An illustration of robot's movement mechanism is shown in Fig. \ref{movement}. By controlling when PCBot attaches and detaches from the surface, we can move robots in the desired direction.

Aside from the movement mechanism, a key system for PCBot is the orientation tracking system made of an ambient light sensor covered by a polarizing filter. Because the light source in the environment is also polarized, the light intensity sensed by the ambient light sensor will change with the orientation of the PCBot. Thus, PCBots can track their orientation by feeding the ambient light strength sensed to a threshold-crossing-based software phase lock loop (sPLL).

\subsection{Magnetic field and geometric design}\label{magnetic-field-and-geometric-design}

The core of PCBot design is the actuation mechanism, i.e. the electro-magnetic subsystem, shown in Fig. \ref{magnetsys}. This subsystem includes a magnet that can move up and down, a coil built in the multi-layer PCB, and a steel core to amplify the magnetic field. The chassis of the robot and the table surface serve to restrict the position of the magnet.

The key of the actuator is the bi-stable state of the magnet. A bi-stable magnet means the robot only needs to power the coil when changing the position of the magnet, but not keeping the magnet in a specific state. This dramatically decreases the power consumption of the mechanism compared to using an electromagnet to pull the magnet up in `detached' state, because switching between states takes $\approx1\,\text{ms}$ but the electromagnet would need to be powered on during the entire `detached' state for tens of milliseconds to seconds.

\begin{figure}[htbp]
\centering
\includegraphics[clip,trim=1.5cm 4.7cm 1.5cm 3.8cm,width=\linewidth]{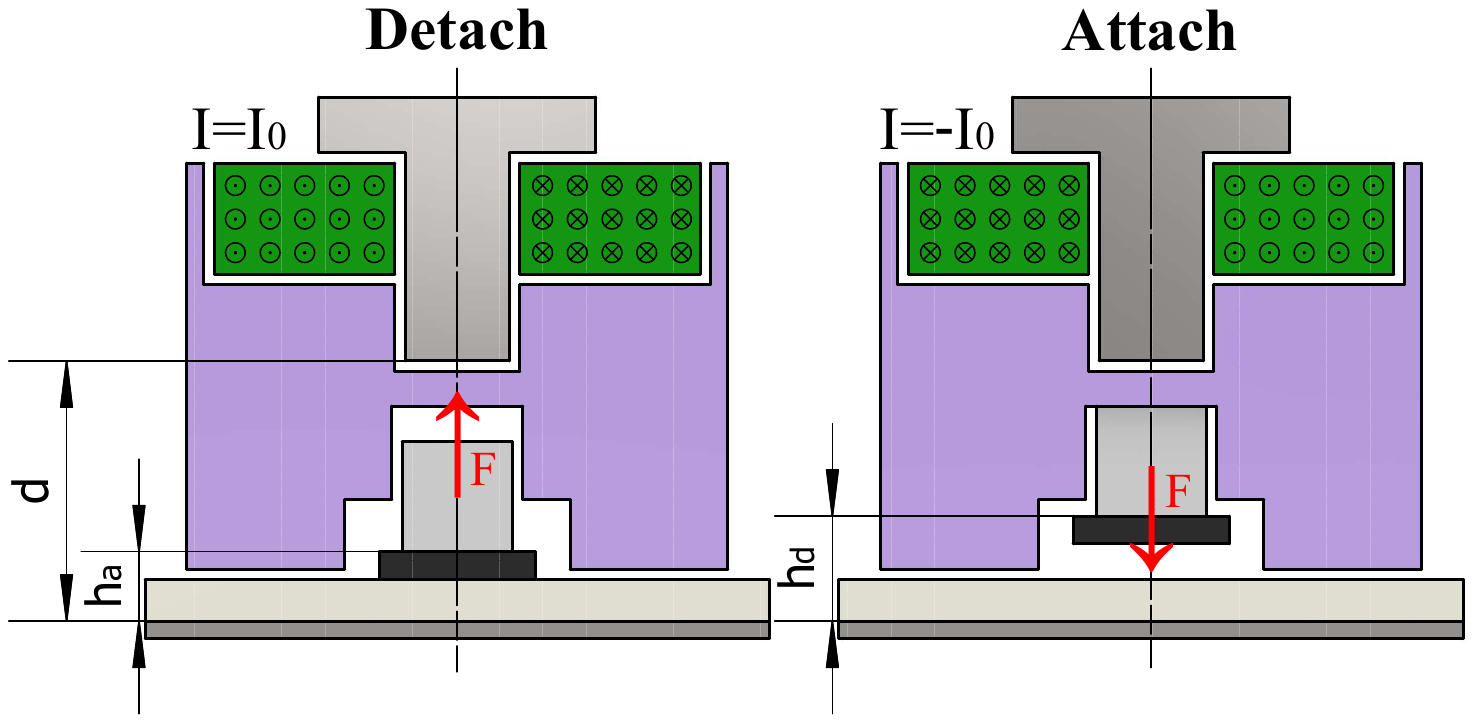}
\caption{A simplified cross-section view of the electro-magnetic subsystem of PCBot. This figure shows how PCBot switches between the `attached' and `detached' states. When detaching, PCBot drives a current $I_0$ through the coil and generates an upward magnetic force $F(d,h_a,I_0)>0$ on the magnet that pulls it upwards. When attaching, PCBot drives a current  $-I_0$ through the coil and generates a downward magnetic force $F(d,h_d,-I_0)<0$ that pushes the magnet downwards.}
\vspace{-3mm}
\label{magnetmov}
\end{figure}

There are two necessary requirements for designing this system: first, that the magnet should be stable in either state when no current flows through the coil, second that the magnet should be able to switch between the states when current flows through the coil, as shown in Fig. \ref{magnetmov}.

We first simplify the model and only consider the three most important and controllable parameters: the distance between the steel core and the steel table $d$, the distance between the magnet and the steel table $h$, and the current through the coil $I$. Other geometric parameters have a limited impact on the result and could be fine-tuned afterwards. The vertical force exerted on the magnet $F(d,h,I)$ could be expressed as a function of $d, h, I$ and computed using commercial finite element methods (FEM) software. Using this notation, our design requirements can be formally written as: 

\begin{equation}
\left\{
\begin{aligned}
F(d,h_{a},0)&<0\\
F(d,h_{d},0)&>0\\
F(d,h,I_0)&>0\ \ (h\in[h_{a},h_{d}])\\
F(d,h,-I_0)&<0\ \ (h\in[h_{a},h_{d}])
\end{aligned}
\right.
\label{eqn2}
\end{equation}

\noindent where $h_{a}, h_{d} $ are the position of the magnet in `attached' and `detached' state respectively, and $I_0$ is the current through the coil when H-bridge is activated. When there is no current through the coil, and $d$ is held constant, as $h$ increases, the attraction force between the steel core and the magnet increases but the attraction force between the steel table and the magnet decreases, thus $F(d,h,0)$ is monotonically increasing with $h$. We can find the only equilibrium point $h=h_{eq}$ that satisfies

\begin{equation}
F(d,h_{eq},0)=0
\end{equation}

\noindent Similarly, we can define $h_{eq}(I)$ as the equilibrium point closest to $h_{eq}$ when current $I$ flows through the coil, i.e. 

\begin{equation}
F(d,h_{eq}(I),I)=0
\label{eqn4}
\end{equation}

\begin{figure}[htbp]
\centering
\includegraphics[width=0.9\linewidth]{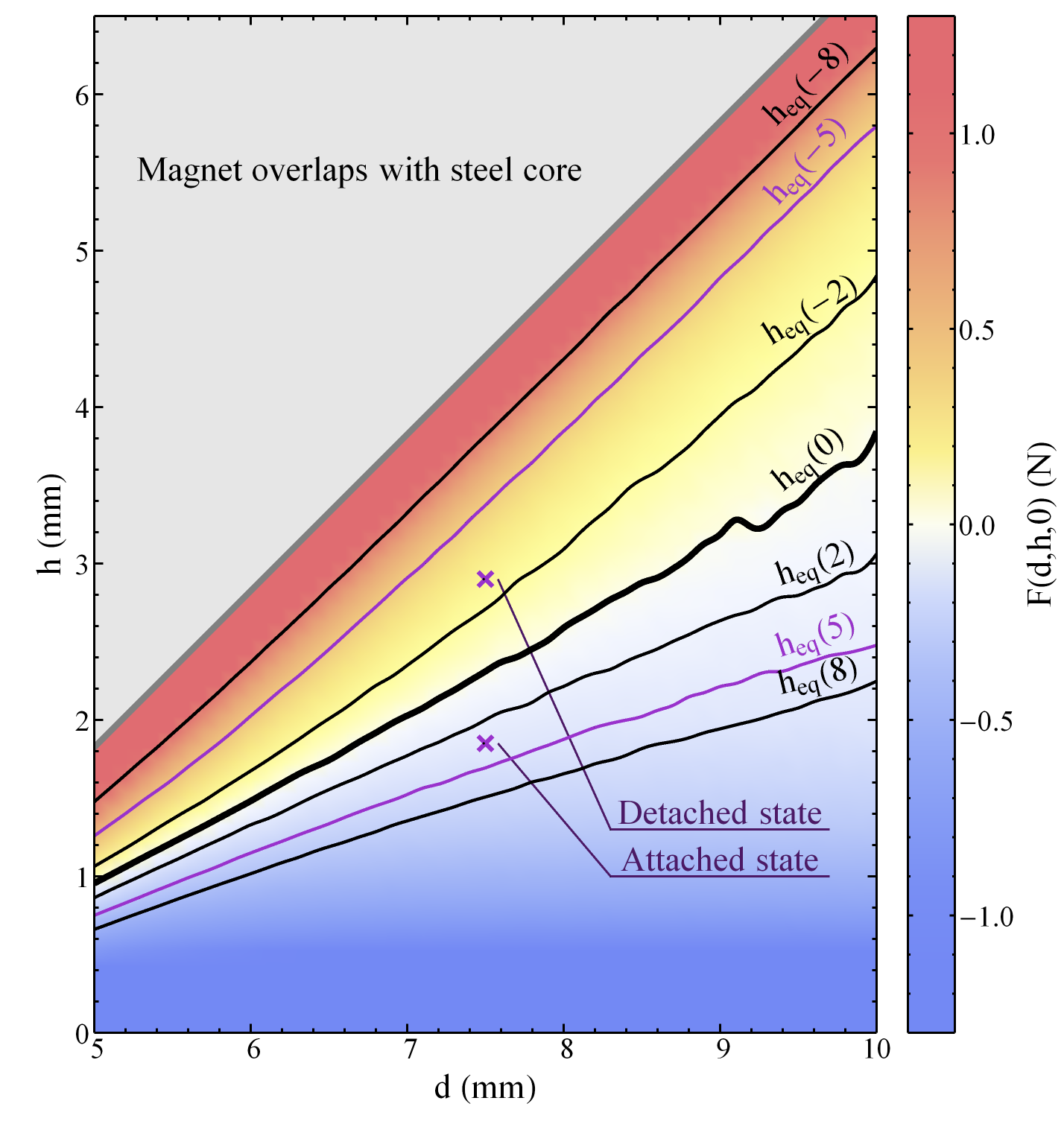}
\caption{Plot of equilibrium point's position $h_{eq}(I)$ under various coil current $I$ and robot geometry parameter $d$. Color in the background shows the force on magnet $F(d,h,0)$. The parameters we chose are $I_0=5\,\text{A}, d=7.5\,\text{mm},h_{a}=1.85\,\text{mm},h_{d}=2.90\,\text{mm}$, as marked purple in the figure. The two crosses, which represent the chosen geometry, are between the two purple lines, with at least $0.1\,\text{mm}$ margin.  This means the coil can generate more than enough force to move the magnet up and down to switch between states.}
\vspace{-3mm}
\label{magnetequ}
\end{figure}

\noindent For example, the plot for $h_{eq}(I)$ and $F(d,h,0)$ with geometric configuration we eventually used is shown in Fig. \ref{magnetequ}. Using these definitions, we can simplify (\ref{eqn2})\textasciitilde{}(\ref{eqn4}) to:

\begin{equation}
h_{eq}(I_0)<h_{a}<h_{eq}(0)<h_{d}<h_{eq}(-I_0)
\label{eqn5}
\end{equation}

When designing the PCBot, we had to choose a proper $I_0,d,h_a$ and $h_d$ that satisfies (\ref{eqn5}). In addition, we wanted to: minimize the power consumption, maximize the friction force between the magnet and the table in `attached' state to prevent slipping, and also make sure that the robot design is robust to manufacturing variability. To minimize power consumption we should try to minimize $I_0$. With the same friction coefficient between the magnet and the table, the magnetic force on the magnet in `attached' state determines the maximum friction force, so we should try to maximize $-F(d,h_a,0)$. Furthermore, manufacturing error will make geometric parameters $d,h_a,h_d$ vary, so it is also important to leave enough margin so (\ref{eqn5}) can still hold with variation in parameters. 

The first thing we chose is the current we are willing to drive through the coil, which is limited by the electronics. In this case $I_0=5\,\text{A}$. Ideally, we would like to increase the force on the magnet to increase friction, which means choosing a point $\{d,h\}$ that is as far left as possible on the equilibrium line $h_{eq}(-I_0)$. However, on the left side of the graph, manufacturing error can result in a large change in the required current to drive the magnet because equilibrium lines are tightly packed, so to compensate for that, we chose a point that is more to the right to match our manufacturing error. This result in the following parameters: $I_0=5\,\text{A}, d=7.5\,\text{mm},h_{a}=1.85\,\text{mm},h_{d}=2.90\,\text{mm}$.

\subsection{Mechanical design}\label{mechanical-design}

\begin{figure}[htbp]
\centering
\includegraphics[clip,trim=2cm 4cm 2cm 4cm,width=\linewidth]{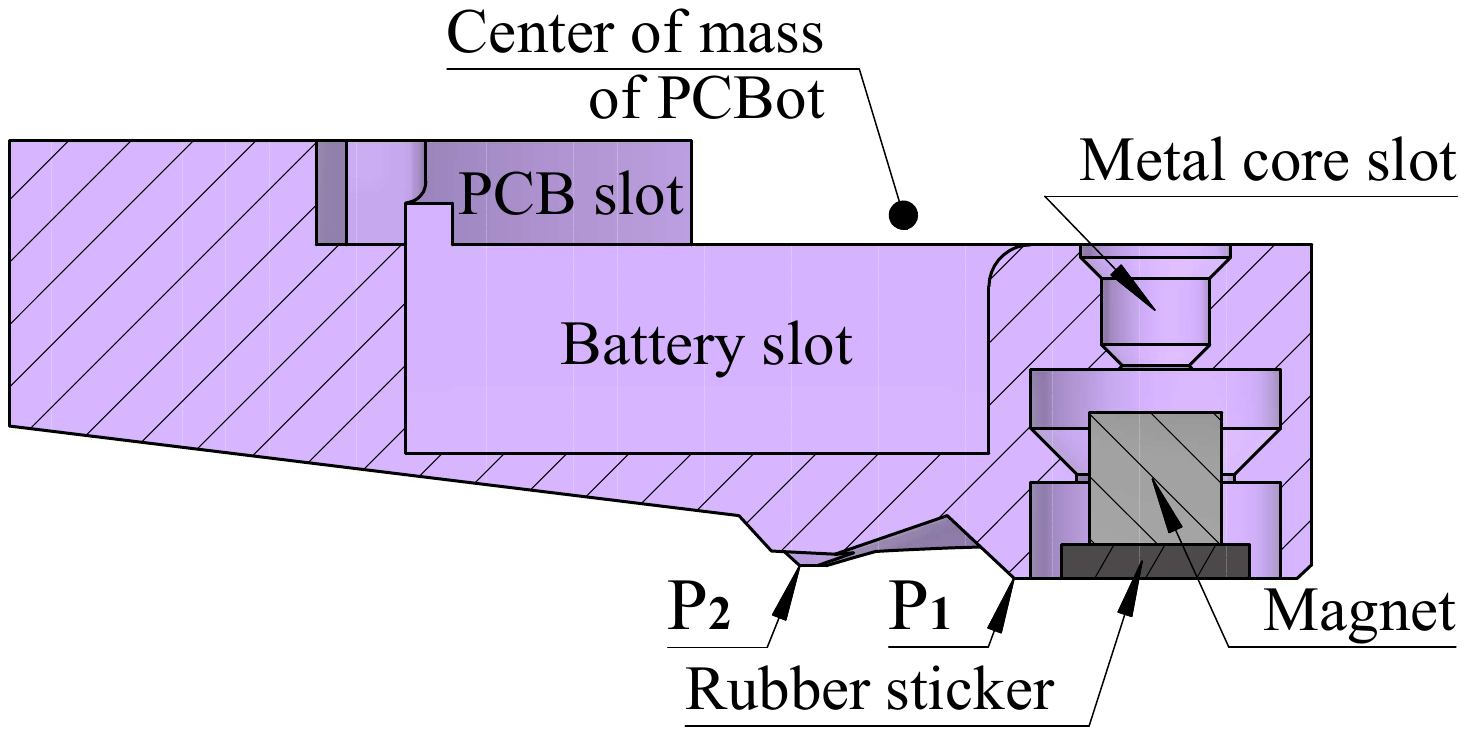}
\caption{Cross-section of robot's 3D printed chassis and the magnet.}
\vspace{-2mm}
\label{mechanical}
\end{figure}

In the mechanical design process, we were mainly concerned with the chassis's shape, the overall mass distribution, and the forces on the magnet. An abstract drawing of the mechanical system is shown in Fig. \ref{mechanical}.

We had three major goals for mechanical design of the system:

\begin{enumerate}
\item
  PCBot must slip along the surface in the `detached' state.
\item
  PCBot must stick to the surface in the `attached' state.
\item
  PCBot must stably rotate around the magnet in the `attached' state.
\end{enumerate}

Goals 1 and 2 mean we had to differentiate the friction force in `detached' and `attached' states as much as possible, so we chose Ultimaker Tough PLA material to 3D print the chassis, moisture-resistant HDPE as the orbital shake table's surface coating, and Neoprene rubber with criss-cross texture as the rubber sticker on the magnet. The static and kinetic friction coefficient between the chassis and the table is around $0.12$ but the kinetic friction coefficient between the rubber and table is around $0.45$ and the static friction coefficient is as high as $0.67$. With this combination of material, the force from kinetic friction in the `attached' state is around $73\,\text{mN}$, but merely $29\,\text{mN}$ in the `detached' state.


Goal 3 means we had to design the chassis so that the torque of the driving force (inertial force in the table's co-moving frame) should be maximized while the torque of friction force acting on the chassis when spinning around the magnet should be minimized. So we designed a special chassis with the cross intersection as shown in Fig. \ref{mechanical}. With this chassis, the total mass of the robot is $18.0\,\text{g}$ and the center of mass is $5.2\,\text{mm}$ away from the robot's axis of rotation. The mass distribution is asymmetric enough so the driving force from the table's movement has enough torque to combat friction in `attached' state. At the same time, the mass of the robot is still small enough so that it won't slip as it spins in the `attached' state. Concerning the torque of the friction force, the chassis has two extrusions on the bottom that will contact the surface: a ring close to the rotation center that contacts the surface at $P_1$, and a line-shaped extrusion further away that contacts the surface at $P_2$. $P_1$ and $P_2$ situate on either side of the center of mass and will contact the surface at all times. This design minimized the torque of friction force in both the `attached' and `detached' states. In comparison, if the chassis is a cylinder shell, in the `attached' state the friction force would act on the boundary of the chassis, making the arm of friction force the radius of the chassis. This is approximately four times longer than that of the current design, in which the arm of friction force is approximately halfway between $P_1$ and $P_2$.

\subsection{Coil design}\label{coil-design}

One of the most significant limitations for small robots is battery life. For PCBot, the power consumption when the actuation mechanism is not activated is $18\,\text{mW}$, which is mainly consumed by the microcontroller. However, if the robot moves once every second, the average power consumption of the actuation mechanism could go above $500\,\text{mW}$, making it the limiting factor for battery life.

The main reason why the actuation mechanism is so power consuming is because we used coils built using a multi-layer PCB to generate magnetic field, so the number of rounds of the coil is a magnitude lower and the resistance is a magnitude larger compared to conventional coils built using regular wires. A viable way to reduce the power consumption of the actuation mechanism without changing the robot's electro-magnetic and mechanical design is to optimize the PCB coil and make it more energy-efficient at moving the magnet. In the following discussion, we will present an abstracted model that ignores manufacturing constraints like vias and component placements to provide an analytical analysis. The PCB coil is built on a multi-layer board, but there is little difference between layers, so we will just analyze a single-layer PCB coil instead.

\begin{figure}[htbp]
\centering
\includegraphics[width=\linewidth]{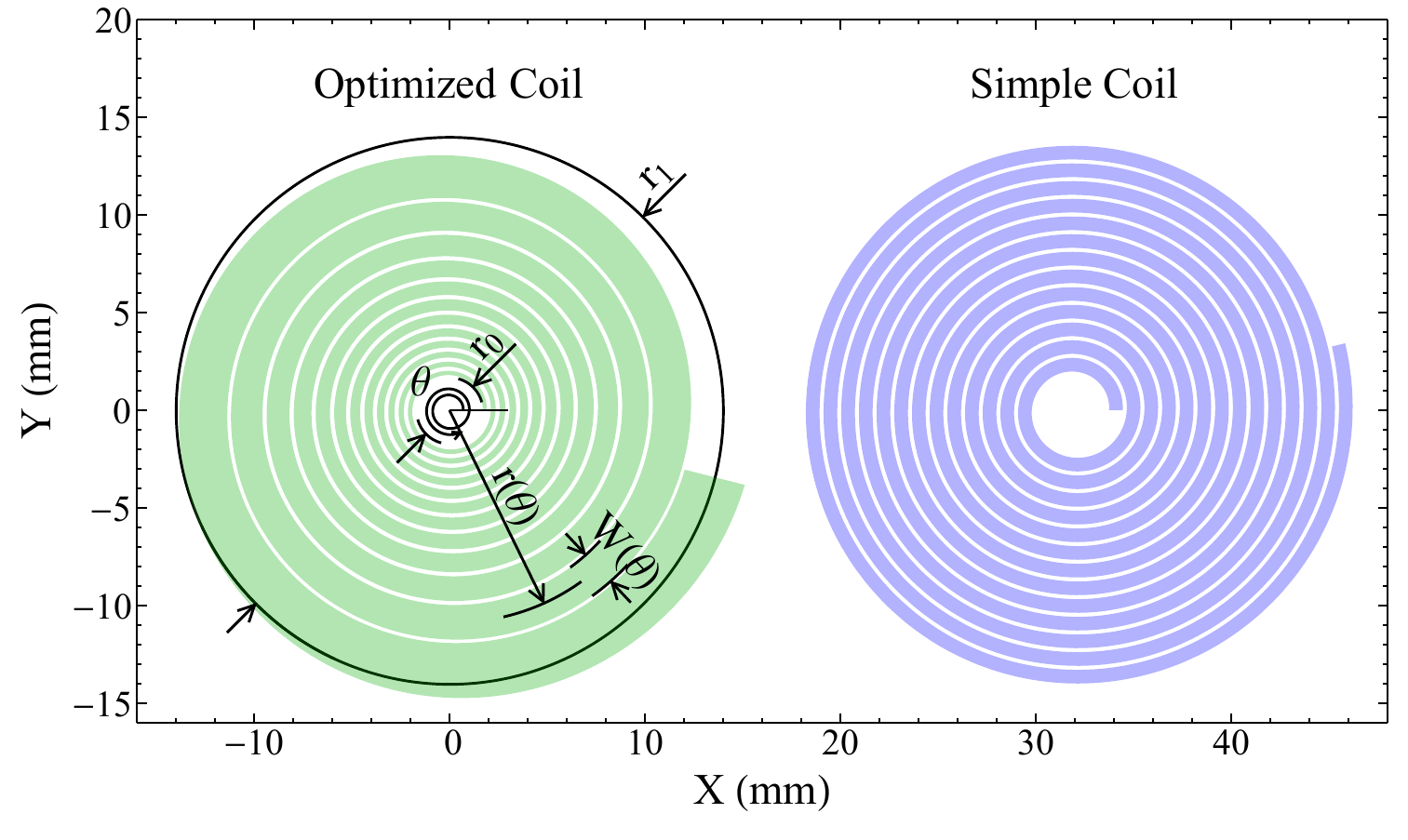}
\caption{The coil's geometric structure for the optimized coil design (left) and simple coil (right).}
\label{coilillus}
\end{figure}

A single-layer PCB coil takes the form of a spiral, as shown in Fig. \ref{coilillus}. The inner and outer radius $r_0, r_1$ of the coil is determined by the size of the steel core and the size of the robot, and the minimum gap between two consecutive rounds of coil $w$ is determined by the PCB manufacturer (usually $0.2\,\text{mm}$ for PCBs with $2\,\text{oz}$ copper weight).

We can describe the PCB coil using function $r(\theta)$ which is the equation of the coil's center line in polar coordinates and $W(\theta)$ which describes the coil's width. We assume that both the radius and the width of the coil change slowly, i.e.

\begin{eqnarray}
r''(\theta) \ll r'(\theta) \ll r\\
W'(\theta) \ll W
\label{coil_assumption}
\end{eqnarray}

Consider the distance $D$ between the center line of two consecutive rounds of coil, we get the relationship between coil's radius $r(\theta)$ and coil's width $W(\theta)$ under assumption (\ref{coil_assumption}).

\begin{equation}
r(\theta+2\pi)-r(\theta)=D=\frac{W(\theta)+W(\theta+2\pi)}{2}+w
\label{coilrw}
\end{equation}

\noindent With assumption (\ref{coil_assumption}), we have $r(\theta+2\pi)-r(\theta)=2\pi r'(\theta+\pi)$ and $W(\theta)+W(\theta+2\pi)=2W(\theta+\pi)$. By applying this to (\ref{coilrw}), we can express $W(\theta)$ using $r(\theta)$ as

\begin{equation}
W(\theta)=\frac{r'(\theta)}{2\pi}-w
\label{coil_width}
\end{equation}

\noindent Now we can describe a coil using only $r(\theta)$. In the next step, we proceed to formulate the magnetic force on magnet $F(r(\theta),I)$ and the power consumption of the coil $P(r(\theta),I)$ with arbitrary coil configuration $r(\theta)$ and coil current $I$ that satisfy (\ref{coil_assumption}).



For the magnetic force on the magnet, because the force generated by the coil is much weaker than the magnetic field of the magnet itself, and all the materials in the system have constant differential permissively under our operating condition, $F(r(\theta),I)$ is almost proportional to the current $I$, and can be written as:

\begin{equation}
F(r(\theta),I)=F(r(\theta)) I
\end{equation}

\noindent where $F(r(\theta))$ can be determined using FEM simulations.

For the power consumption of the coil, because we usually need to actuate the coil for around $1\,\text{ms}$ to move the magnet, and the characteristic time for the LR circuit composed by the power source and the coil $t_{LR}$ has $t_{LR}\approx10^1\,\text{us}\ll 1\,\text{ms}$, Joule's law could be used to calculate the power consumption:

\begin{equation}
P(r(\theta),I)=I^2(R_{ext}+R_{coil})
\end{equation}

\noindent where $R_{ext}$ is the internal resistance of the actuation circuit, including the internal resistance of the power source (ESR of the capacitor in our case) and the on state resistance of the H-bridge, and

\begin{equation}
\begin{split}
R_{coil}&=\int_{r_0}^{r_1} \rho_s \frac{r(\theta)\,\dd \theta}{W(\theta)}\\
&=\int_{r_0}^{r_1} \rho_s \frac{r(\theta)}{r'(\theta)(\frac{r'(\theta)}{2\pi}-w)} \dd r
\end{split}
\end{equation}

\noindent where $\rho_s$ is the surface resistivity. We can observe that $P(r(\theta),I)/F(r(\theta),I)^2$ is a constant regardless of current $I$ and is only related to the coil design $r(\theta)$. Thus the optimal coil configuration can be obtained by numerically minimizing

\begin{equation}
\frac{P(r(\theta),I)}{F(r(\theta),I)^2}=\frac{R_{ext}+\rho_s \int_{r_0}^{r_1} \frac{r(\theta)}{r'(\theta)(\frac{r'(\theta)}{2\pi}-w)} \dd r}{F(r(\theta))^2}
\end{equation}

In realistic conditions with $r_0=2\,\text{mm}$, $r_1=14\,\text{mm}$, $R_{ext}=0.25\,\Omega$, $\rho_s=0.25\,\text{m}\Omega$, the optimized coil shown in Fig. \ref{coilillus} can save $40\%$ energy comparing to a simple coil with fixed $0.7\text{mm}$ width.  Furthermore, geometric constraints created by vias and nearby components severely reduced the performance of the simple coil, while the optimized coil was not influenced as much. By using the optimized coil in PCBot, we saved a staggering $70\%$ of energy compared to using a simple coil.

\section{Experiments}\label{experiments}

We conducted several experiments to demonstrate the capability of the PCBot and test its performance. When not specified, the orbital shake table moved at $115\pm 2\,\text{rpm}$ at an orbital radius of $10.7\,\text{mm}$, and the robot would lift the magnet up for $100\,\text{ms}$ in each actuation cycle, resulting in an average displacement of $5\,\text{mm}$. Robots are autonomous, and there is no position feedback.

We recorded the movement of the table and the robot using an overhead calibrated camera, and then track the fiducial trackers on the table and the robot to obtain their translation and rotation.

\subsection{Straight line movement}\label{straight-line-movement}

\begin{figure}[htbp]
\centering
\includegraphics[clip,trim=0cm 0.6cm 0cm 0.15cm, width=\linewidth]{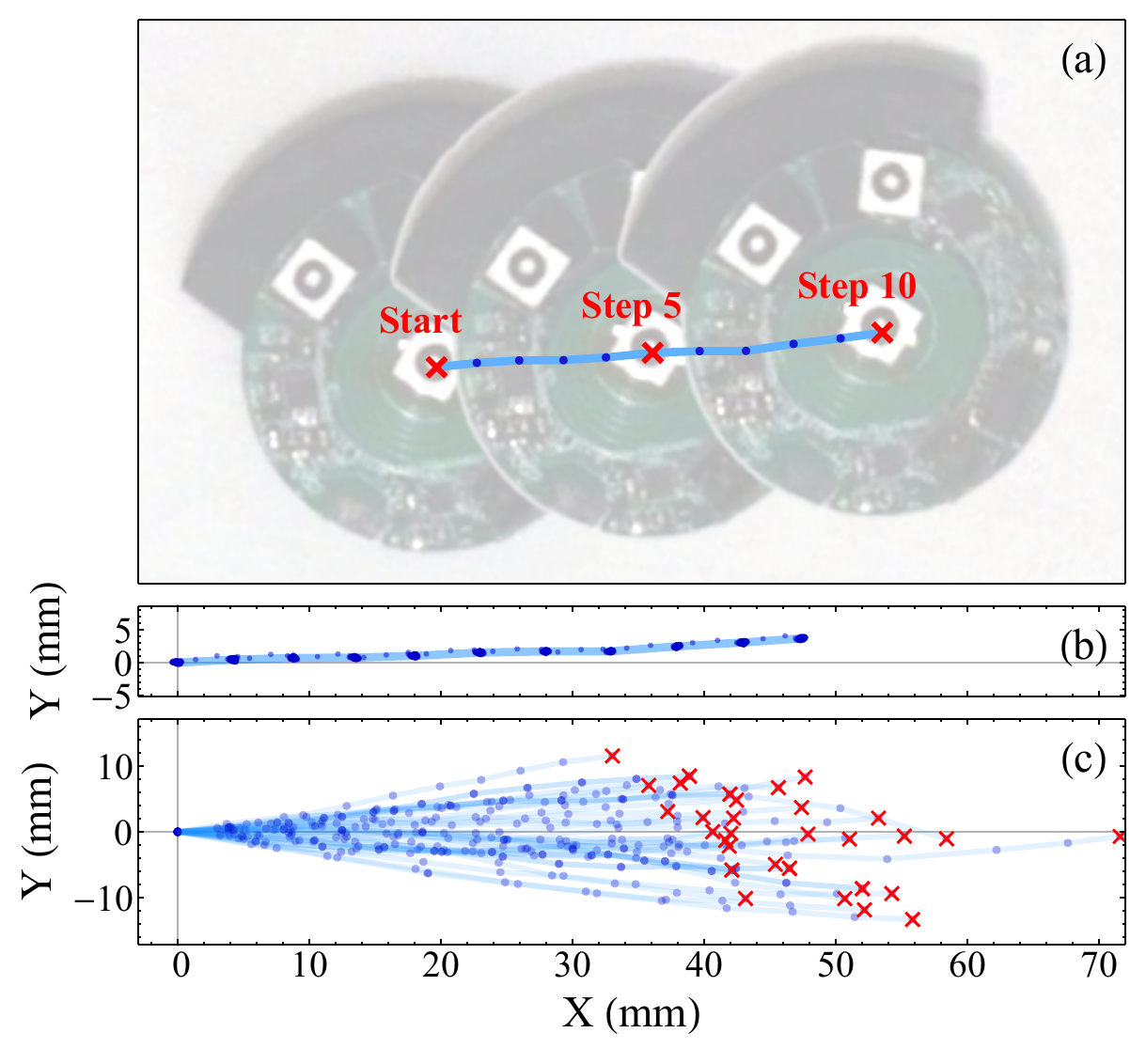}
\caption{Experimental results for straight line movement. Subfigure (a) is a composition of PCBot's configuration in the initial state, and after moving 5 and 10 actuation cycles in a typical experiment. Red crosses and blue dots show the magnet's starting position in each actuation cycle, and light blue lines connecting these dots show the path robot takes. Subfigure (b) is a more detailed view of magnet's position in the experiment corresponding to Subfigure (a). Subfigure (c) shows the magnet's path in all 30 experiments. The red cross in this figure marks the ending point in each trial.}
\label{linemove}
\end{figure}

We first conducted 36 experiments where a PCBot is programmed to walk in a straight line following the X-axis for 10 actuation cycles ($\approx 47\,\text{mm}$) to test the reliability of movement mechanism. Experiments are carried out in different places on the table's surface to test its sensitivity to the surface quality. The results are shown in Fig. \ref{linemove}.

\begin{figure}[htbp]
\centering
\includegraphics[clip,trim=0cm 0.6cm 0cm 0.15cm,width=\linewidth]{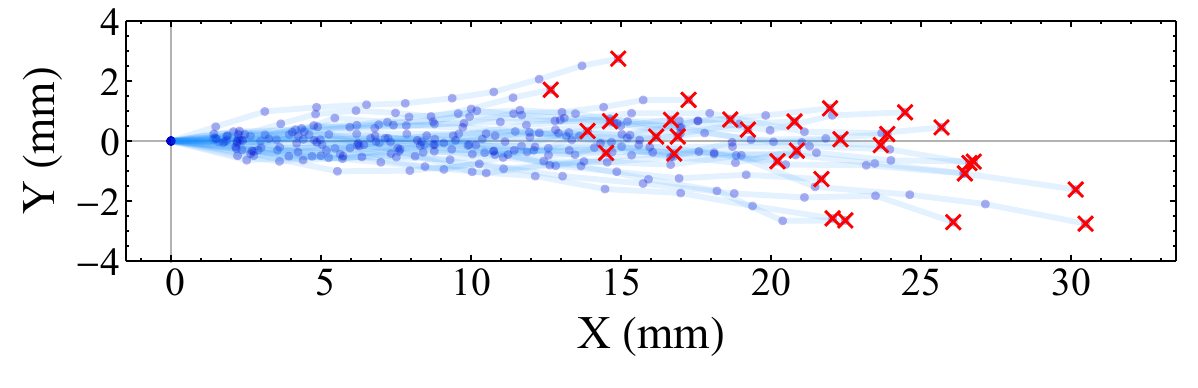}
\caption{Experimental results for straight line movement when magnet are lifted for $50\,\text{ms}$ in each actuation cycle. Notations are the same as Fig. \ref{linemove}.(c)}
\label{linemove2}
\end{figure}

For comparison, we also did 30 similar experiments but with slightly different parameters and half the magnet up time ($50\,\text{ms}$ in each actuation cycle). The results are shown in Fig. \ref{linemove2}.

We observed that the PCBot would not slip in `attached' state, but would rapidly move in the desired direction in `detached' state. When the magnet is lifted for $100\,\text{ms}$ in each actuation cycle, the average displacement per 10 actuation cycles is $47\,\text{mm}$ and the standard deviation for the displacement's direction and distance are $8^\circ$ and $7\,\text{mm}$ respectively. When the magnet is lifted for $50\,\text{ms}$ in each actuation cycle, the average displacement per 10 actuation cycles is $21\,\text{mm}$ and the standard deviation for the displacement's direction and distance are $4^\circ$ and $5\,\text{mm}$ respectively. The accuracy of direction increases while the accuracy of distance decreases when the magnet is lifted for shorter period of time. Accuracy in both cases is sufficient as long as there is external feedback to correct the robot's course.

By analyzing the correlation between the robot's movement and position, we discovered that the major cause for its non-uniform displacement is the difference in friction coefficient across the HDPE sheet used. The moving direction is related to the phase lag between the rotation of the PCBot and the rotation of the table, and the moving distance is related to the acceleration of PCBot when slipping on the table. Both the phase lag and the acceleration depends on the friction coefficient between the table and the chassis. This problem could be reduced by switching to a sheet with better uniformity across the surface or reducing step size by reducing the time in the `detached' state in each actuation cycle. Other factors that might contribute to the non-uniformity are the slight variation in table rotation speed, imprecision in leveling of the table, and bumps on the table surface.

\subsection{Path following}\label{path-following}

\begin{figure}[htbp]
\centering
\includegraphics[clip,trim=0cm 0.5cm 0cm 0.3cm,width=\linewidth]{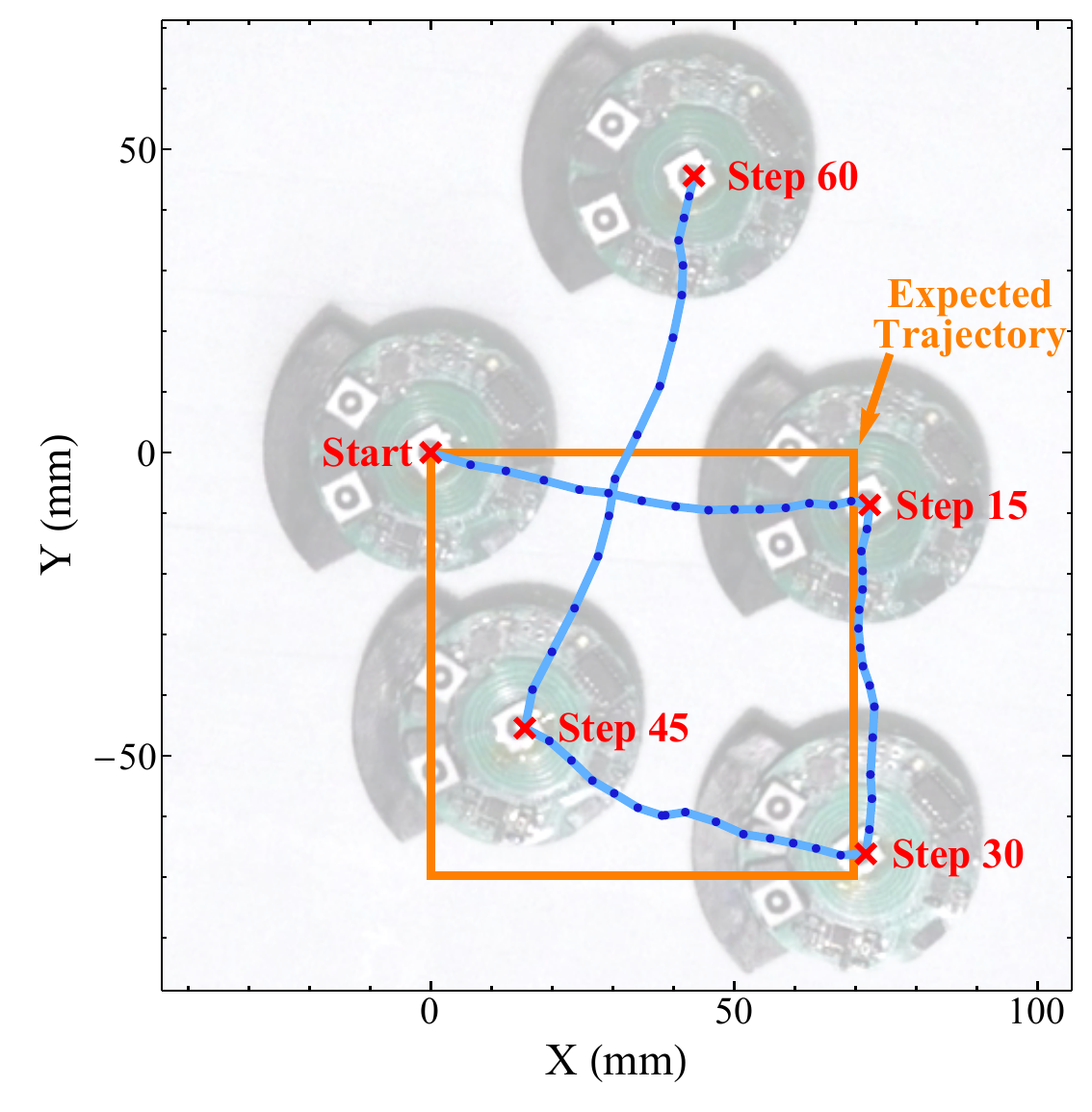}
\caption{Experimental results for path following experiment. This figure shows the composition of PCBot's position in the initial state and after moving every 15 actuation cycles. Red crosses and blue dots shows the magnet's starting position in each actuation cycle, light blue lines connecting these dots shows the path robot takes, and the yellow path shows the expected trajectory of PCBot. }
\vspace{-2.5mm}
\label{pathmove}
\end{figure}

We would also like to demonstrate the PCBot's path following capability, so we programmed a PCBot to move in a rectangular pattern where each edge is 15 actuation cycles long ($\approx 70\,\text{mm}$). The results are shown in Fig. \ref{pathmove}.

As shown in Fig. \ref{pathmove}, the PCBot could follow the path we gave. However, because PCBot does not have position feedback, the final position at step 60 is not aligned with the starting position. Their movement direction is relatively accurate, but their movement speed was not as consistent.  Again, this is likely due to the variability of friction across HDPE sheet.

\subsection{Power consumption analysis}\label{power-consumption-analysis}

As suggested in section \ref{coil-design}, a general concern for small robots is battery life. In our case, the total distance PCBot can move is directly related to the number of actuations the on-board battery can support.  Thus, it is important to know how much energy PCBot consumes in each actuation cycle and how many actuation cycles can the battery support. In our experiments, actuation time was set to $1000\,\text{us}$ for detaching and attaching. On average, the power consumption when the movement mechanism was not activated was $18\,\text{mW}$, and each actuation cycle cost $68\,\text{mJ}$ of energy. If we only consider the power consumption of the actuation mechanism, the $70\,\text{mAh}$ on-board Li-ion battery allows the PCBot to move for about 14,000 actuation cycles. Under current operating conditions where the PCBot moves an average of $4.7\,\text{mm}$ per actuation cycle, PCBots can move for about $66\,\text{m}$ before the battery depletes. However, the distance PCBots can move in each actuation cycle can be further increased if we increase the time in `detached' state. When PCBots stay in the `detached' state for more than $200\,\text{ms}$ in each actuation cycle, they can move more than $10\,\text{mm}$ per step. This means that PCBots could travel for more than $140\,\text{m}$ before the battery depletes. Considering that PCBots are less than $5\,\text{cm}$ in size, this is a significant distance.

\section{Conclusion}\label{conclusion}

In this paper, we presented PCBot, a minimalist robot for swarm applications, explained its theory of operation, optimized its performance, and demonstrated its movement capability in real experiments. The design based on external actuation and novel bi-stable solenoid actuator built using PCB-based coil allows PCBots to possess basic control and precise movement capability with merely five major components.

PCBot only demonstrates its ability to move at present and lacks the inter-robot sensing and communication needed for a complete swarm robot.  In future work, we will outfit them with IR transceivers to allow them to perform inter-robot communication and bearing/distance sensing, similar to \cite{gyongyosi2017low, saloutos2019spinbot}. It is worth noting that all these could be done by simply adding more circuitry to the PCB and should not involve adding more complexity to the assembly. With these future additions, PCBot would be a very simple, mass-manufacturable, yet powerful robot, making it a great candidate as testbed for a variety of swarm algorithms.

\printbibliography

@inproceedings{rubenstein2013collective,
  title={Collective transport of complex objects by simple robots: theory and experiments},
  author={Rubenstein, Michael and Cabrera, Adrian and Werfel, Justin and Habibi, Golnaz and McLurkin, James and Nagpal, Radhika},
  booktitle={AAMAS 2013},
  pages={47--54},
  year={2013}
}

@inproceedings{chiu2009tentacles,
  title={Tentacles: Self-configuring robotic radio networks in unknown environments},
  author={Chiu, Harris Chi Ho and Ryu, Bo and Zhu, Hua and Szekely, Pedro and Maheswaran, Rajiv and Rogers, Craig and Galstyan, Aram and Salemi, Behnam and Rubenstein, Mike and Shen, Wei-Min},
  booktitle={IROS},
  pages={1383--1388},
  year={2009},
  organization={IEEE}
}

@incollection{gauci2018programmable,
  title={Programmable self-disassembly for shape formation in large-scale robot collectives},
  author={Gauci, Melvin and Nagpal, Radhika and Rubenstein, Michael},
  booktitle={DARS},
  pages={573--586},
  year={2018},
  publisher={Springer}
}

@article{gross2009towards,
  title={Towards group transport by swarms of robots},
  author={Gro{\ss}, Roderich and Dorigo, Marco},
  journal={International Journal of Bio-Inspired Computation},
  volume={1},
  number={1-2},
  pages={1--13},
  year={2009},
  publisher={Inderscience Publishers}
}

@inproceedings{ceron2019towards,
  title={Towards a Scalable, Self-Reconfigurable Robot with Compliant Modules},
  author={Ceron, Steven and Horowitz, Logan and Wilson, Nialah and Chen, Claire and Kim, Daniel and Petersen, Kirstin},
  booktitle={IEEE MRS 2019},
  pages={47--49},
  year={2019}
}

@inproceedings{gyongyosi2017low,
  title={Low cost sensing and communication system for rotor-craft},
  author={Gyongyosi, Marc and Daley, Alexander and Resnick, Blake and Rubenstein, Michael},
  booktitle={IROS 2017},
  pages={},
  year={}
}

@inproceedings{swissler2018fireant,
  title={FireAnt: A modular robot with full-body continuous docks},
  author={Swissler, Petras and Rubenstein, Michael},
  booktitle={ICRA 2018},
  pages={6812--6817},
  year={2018}
}

@article{wurman2008coordinating,
  title={Coordinating hundreds of cooperative, autonomous vehicles in warehouses},
  author={Wurman, Peter R and D'Andrea, Raffaello and Mountz, Mick},
  journal={AI magazine},
  volume={29},
  number={1},
  pages={9--9},
  year={2008}
}

@article{yim2007modular,
  title={Modular self-reconfigurable robot systems [grand challenges of robotics]},
  author={Yim, Mark and Shen, Wei-Min and Salemi, Behnam and Rus, Daniela and Moll, Mark and Lipson, Hod and Klavins, Eric and Chirikjian, Gregory S},
  journal={IEEE RAM},
  volume={14},
  number={1},
  pages={43--52},
  year={2007},
  publisher={IEEE}
}

@inproceedings{gilpin2010robot,
  title={Robot pebbles: One centimeter modules for programmable matter through self-disassembly},
  author={Gilpin, Kyle and Knaian, Ara and Rus, Daniela},
  booktitle={ICRA 2010},
  pages={2485--2492},
  year={2010}
}

@article{fox2006distributed,
  title={Distributed multirobot exploration and mapping},
  author={Fox, Dieter and Ko, Jonathan and Konolige, Kurt and Limketkai, Benson and Schulz, Dirk and Stewart, Benjamin},
  journal={Proceedings of the IEEE},
  volume={94},
  number={7},
  pages={1325--1339},
  year={2006},
  publisher={IEEE}
}

@article{wilson2020scalable,
  title={Scalable and robust fabrication, operation, and control of compliant modular robots},
  author={Wilson, Nialah Jenae and Ceron, Steven and Horowitz, Logan and Petersen, Kirstin},
  journal={Frontiers in Robotics and AI},
  volume={7},
  pages={44},
  year={2020},
  publisher={Frontiers}
}

@INBOOK{6301066,
  author={Durrant-Whyte, Hugh and Roy, Nicholas and Abbeel, Pieter},
  booktitle={RSS 2012}, 
  title={Construction of Cubic Structures with Quadrotor Teams}, 
  year={2012},
  volume={},
  number={},
  pages={177-184},
  doi={}}

@article{GREGORY2020e00105,
  title={microUSV: A low-cost platform for indoor marine swarm robotics research},
  author={Gregory, Calvin and Vardy, Andrew},
  journal={HardwareX},
  volume={7},
  pages={e00105},
  year={2020},
  publisher={Elsevier}
}

@inproceedings{nisser2016feedback,
  title={Feedback-controlled self-folding of autonomous robot collectives},
  author={Nisser, Martin EW and Felton, Samuel M and Tolley, Michael T and Rubenstein, Michael and Wood, Robert J},
  booktitle={IROS 2016},
  pages={1254--1261},
  year={2016}
}

@article{felton2014method,
  title={A method for building self-folding machines},
  author={Felton, Samuel and Tolley, Michael and Demaine, Erik and Rus, Daniela and Wood, Robert},
  journal={Science},
  
  pages={644--646},
  year={2014},
  publisher={American Association for the Advancement of Science}
}

@inproceedings{mondada2009puck,
  title={The e-puck, a robot designed for education in engineering},
  author={Mondada, Francesco and Bonani, Michael and Raemy, Xavier and Pugh, James and Cianci, Christopher and Klaptocz, Adam and Magnenat, Stephane and Zufferey, Jean-Christophe and Floreano, Dario and Martinoli, Alcherio},
  booktitle={ICARSC 2009},
  volume={1},
  number={CONF},
  pages={59--65},
  year={2009},
  organization={IPCB: Instituto Polit{\'e}cnico de Castelo Branco}
}

@inproceedings{mondada2003swarm,
  title={Swarm-bot: From concept to implementation},
  author={Mondada, Francesco and Guignard, Andr{\'e} and Bonani, Michael and Bar, Daniel and Lauria, Michel and Floreano, Dario},
  booktitle={IROS 2003},
  volume={2},
  pages={1626--1631},
  year={2003}
}

@article{rubenstein2014programmable,
  title={Programmable self-assembly in a thousand-robot swarm},
  author={Rubenstein, Michael and Cornejo, Alejandro and Nagpal, Radhika},
  journal={Science},
 
  pages={795--799},
  year={2014},
  publisher={American Association for the Advancement of Science}
}

@book{fortunic2017design,
  title={Design and implementation of a bristle bot swarm system},
  author={Fortuni{\'c}, Juan Edmundo Pozo},
  year={2017},
  publisher={Pontificia Universidad Catolica del Peru-CENTRUM Catolica (Peru)}
}

@INPROCEEDINGS{8206295,
  author={Weston-Dawkes, William P. and Ong, Aaron C. and Majit, Mohamad Ramzi Abdul and Joseph, Francis and Tolley, Michael T.},
  booktitle={IROS 2017}, 
  title={Towards rapid mechanical customization of cm-scale self-folding agents}, 
  year={2017},
  volume={},
  number={},
  pages={4312-4318}}

@INPROCEEDINGS{4107855,
  author={Woern, Heinz and Szymanski, Marc and Seyfried, Joerg},
  booktitle={RO-MAN 2006}, 
  title={The I-SWARM project}, 
  year={2006},
  volume={},
  number={},
  pages={492-496}}

@ARTICLE{10.3389/frobt.2017.00055,
  
AUTHOR={Nemitz, Markus P. and Sayed, Mohammed E. and Mamish, John and Ferrer, Gonzalo and Teng, Lijun and McKenzie, Ross M. and Hero, Alfred O. and Olson, Edwin and Stokes, Adam A.},   
	 
TITLE={HoverBots: Precise Locomotion Using Robots That Are Designed for Manufacturability},      
	
JOURNAL={Frontiers in Robotics and AI},      
	
VOLUME={4},      
	
YEAR={2017},
}

@inproceedings{piccoli2017piccolissimo,
  title={Piccolissimo: The smallest micro aerial vehicle},
  author={Piccoli, Matthew and Yim, Mark},
  booktitle={ICRA 2017},
  pages={3328--3333},
  year={2017}
}

@incollection{saloutos2019spinbot,
  title={SpinBot: An Autonomous, Externally Actuated Robot for Swarm Applications},
  author={SaLoutos, Andrew and Rubenstein, Michael},
  booktitle={Distributed Autonomous Robotic Systems},
  pages={211--224},
  year={2019},
  publisher={Springer}
}

@inproceedings{wang2016autonomous,
  title={Autonomous mobile robot with independent control and externally driven actuation},
  author={Wang, Hanlin and Rubenstein, Michael},
  booktitle={IROS},
  pages={},
  year={2016}
}

@inproceedings{zhang2016controllable,
  title={A controllable flying vehicle with a single moving part},
  author={Zhang, Weixuan and Mueller, Mark W and D'Andrea, Raffaello},
  booktitle={ICRA 2016},
  pages={3275--3281},
  year={2016}
}

@INPROCEEDINGS{6907226,
  author={Zarrouk, David and Fearing, Ronald S.},
  booktitle={ICRA}, 
  title={1STAR, A one-actuator steerable robot}, 
  year={2014},
  volume={},
  number={},
  pages={ }}

\end{document}